\documentclass[runningheads]{llncs}
\usepackage{times}
\usepackage{soul}
\usepackage{url}
\usepackage[hidelinks]{hyperref}
\usepackage[utf8]{inputenc}
\usepackage{graphicx}
\usepackage{amsmath}

\usepackage{amsthm}
\usepackage{booktabs}
\usepackage{algorithm}
\usepackage{algorithmic}
\usepackage{amssymb}% http://ctan.org/pkg/amssymb
\usepackage{pifont}% http://ctan.org/pkg/pifont
\usepackage{rotating}
\usepackage[T1]{fontenc}

\usepackage{graphicx}
\usepackage{subcaption}
\usepackage{booktabs}
\begin{document}
	\title{Mask Atari for Deep Reinforcement Learning as POMDP Benchmarks}
	%\author{Yang SHAO\inst{1}\orcidID{0000-1111-2222-3333} \and
	%Second Author\inst{2,3}\orcidID{1111-2222-3333-4444} \and
	%Third Author\inst{3}\orcidID{2222--3333-4444-5555}}
	\author{Yang Shao \and
		Quan Kong \and
		Tadayuki Matsumura \and
		Taiki Fuji \and
		Kiyoto Ito \and
		Hiroyuki Mizuno
	}
	\authorrunning{Y.Shao et al.}
	\institute{Hitachi Ltd., Higashi koigakubo 1-280, Kokubunji-shi, Tokyo, Japan
		\email{yang.shao.kn@hitachi.com}}
	%\\
	%ABC Institute, Rupert-Karls-University Heidelberg, Heidelberg, %Germany\\
	\maketitle              % typeset the header of the contribution
	%
	%Deep reinforcement learning,POMDP,Atari,Active information gathering
	
	\begin{abstract}
		We present Mask Atari, a new benchmark to help solve partially observable Markov decision process (POMDP) problems with Deep Reinforcement Learning (DRL)-based approaches. To achieve a simulation environment for the POMDP problems, Mask Atari is constructed based on Atari 2600 games with controllable, moveable, and learnable masks as the observation area for the target agent, especially with the active information gathering (AIG) setting in POMDPs. Given that one does not yet exist, Mask Atari provides a challenging, efficient benchmark for evaluating the methods that focus on the above problem. Moreover, the mask operation is a trial for introducing the receptive field in the human vision system into a simulation environment for an agent, which means the evaluations are not biased from the sensing ability and purely focus on the cognitive performance of the methods when compared with the human baseline. We describe the challenges and features of our benchmark and evaluate several baselines with Mask Atari. %The environment of Mask Atari is available at \url{https://github.com/celarex/Mask-Atari}
	\end{abstract}
	
	\section{Introduction}
	As deep reinforcement learning (DRL) has evolved, ambitious challenges such as multi-player real-time strategy games \cite{vinyals2017starcraft,berner2019dota} and three-dimensional virtual environments \cite{beattie2016deepmind,johnson2016malmo,kempka2016vizdoom} have been proposed. Active Information Gathering (AIG) in partially observable Markov decision process (POMDP) \cite{astrom1965optimal} environments for complex tasks is a critical sub-challenge for achieving them. Availability of simulated domains with an appropriate level of difficulty and variety is critical for measuring and advancing algorithms for the sub-challenge. 
	
	However, directly testing AIG ideas in the challenge POMDP environments mentioned above is too computationally heavy and gets entangled with other challenges, such as three-dimensional object recognition and multi-agent optimization. Despite the fact that obtaining optimal policies for POMDPs is computationally hard \cite{madani2003undecidability}, there have been substantial breakthroughs in research on approximate planning methods for POMDPs. With this progress, a wide range of benchmarks have been proposed to test solvers, from cases with several states (e.g., Tiger, Shuttle, Cheese Maze, Tiger-grid, Hallway, and Hallway2 \cite{littman1995learning}) to cases with millions of states or observations (e.g., Rock Sample \cite{smith2012heuristic} and Laser Tag \cite{somani2013despot}). Several cases with continuous states, observations, and action spaces (e.g., Light Dark and Sub Hunt \cite{sunberg2018online}) were also proposed. Pocman, a POMDP modification of the video game Pacman (as described in \cite{silver2010monte}), was also proposed to show that DRL can handle vision-rich cases with a tremendous number of states. Another vision-rich case is Flickering Atari \cite{hausknecht2015deep}, which was proposed to test the effectiveness of a deep neural network as history memory for the information state approach. It modified the Atari games by obscuring the entire screen with a certain probability at each frame. However, although the modification provided a simulation of sensor noise, the field-of-view limitations were not simulated, and no extra actions were supplied for AIG. 
	Therefore, to solve the grand challenges mentioned above step by step, we require a POMDP benchmark focused on AIG for complex tasks for situations with a field-of-view limitation in particular. 
	%In this work, 
	%we propose a set of Atari-based POMDP benchmark environments that focus on the AIG challenge for complex tasks. By creating a uniformed mask interface and extra action spaces, 
	%we construct a flexible, convenient framework to simulate a field-of-view limitation. In this way, we provide a test bed for research on AIG in complex, objective, vision-rich tasks. Moreover, our procedures can be compared with the human baseline and, thus, obtain an accurate idea of the gap between general agents and human players. 
	%After providing the performance of widely used convolutional neural network (CNN) and recurrent neural network (RNN) models as reference baselines for the benchmark cases, we test two heuristic ideas for AIG to show the opportunities and challenges of the hand-eye coordination required by Mask Atari. 
	The contributions of this benchmark are the following.
	\begin{enumerate}
		\item Proposing a set of Atari-based POMDP benchmark environments that focus on the AIG challenge for complex tasks by creating a uniformed mask interface and extra action spaces. 
		\item Supplying the performance of widely used CNN and RNN models as reference baselines and two heuristic ideas for AIG to show the opportunities and challenges of hand-eye coordination in the benchmark environments. 
	\end{enumerate}
	In this way, we provide a test bed for research on AIG in complex, objective, vision-rich tasks. Moreover, our procedures can be compared with the human baseline and, thus, obtain an accurate idea of the gap between general agents and human players. 
	
	\section{Related Work}
	\subsubsection{POMDP Benchmarks}
	Table \ref{tab:POMDPEnvs} shows major benchmark cases used in past POMDP planning research. \cite{littman1995learning} proposed Tiger-grid, Hallway, and Hallway2 as benchmarks over Tiger, Shuttle, Cheese maze, and other cases in earlier research. \cite{bonet1998solving} proposed Heaven/Hell, a new case that requires AIG. \cite{smith2012heuristic} proposed a new scale expandable Rock Sample environment. Rock Sample also provided an independent AIG-action-simulated field-of-view limitation. \cite{eaton2017robust} considered a field-of-view limitation for the tracking task of UAV. \cite{kurniawati2008sarsop} proposed benchmark cases with over ten thousand states and up to a thousand observations, such as Underwater Navigation, Grasping, Homecare, and Integrated Exploration. The scale of new benchmark cases has increased rapidly with the efficiency of new approximate methods. \cite{somani2013despot} proposed Laser Tag, which has over $1.5\times10^{6}$ observations. \cite{silver2010monte} proposed Pocman and battleship. Pocman has around $10^{56}$ states and 1,024 observations, and battleship has around $10^{18}$ states and 100 actions. In DRL literature, \cite{hausknecht2015deep} tried to use an RNN to encode history for vision-rich POMDP cases and proposed Flickering Atari. They modified the Atari games by obscuring the entire screen with a certain probability at each frame. 
	\begin{table}[tb]
		\caption{List of POMDP Environments. AIG: Active Information Gathering.}
		\centering
		\begin{tabular}{lcccc}
			\hline
			Environment & State & Obs. & Act. & AIG \\
			\hline
			%Tiger & 2 & 2 & 3 & - \\
			%Shuttle & 8 & 5 & 3 & - \\
			%Cheese maze & 11 & 7 & 4 & - \\
			%$4\times4$ grid & 16 & 2 & 4 & - \\
			Tiger, Shuttle, Cheese maze, $4\times4$ grid & <16 & <7 & <4 & - \\
			Heaven/Hell & 20 & 11 & 4 & $\checkmark$ \\ 
			%Tiger-grid & 33 & 17 & 5 & - \\
			%Hallway & 57 & 21 & 5 & - \\
			%Hallway2 & 89 & 17 & 5 & - \\
			%Tag & 870 & 30 & 5 & - \\
			Tiger-grid, Hallway, Hallway2, Tag & <870 & <30 & <5 & - \\
			%RS[5,5] & 801 & 2 & 10 & $\checkmark$ \\
			RS[7,8] & 12545 & 2 & 13 & $\checkmark$ \\
			%Grasping & 1253 & 96 & 6 & - \\
			%Underwater & 2653 & 103 & 6 & - \\
			%Homecare & 5408 & 928 & 9 & - \\
			%Integrated & 15517 & 1015 & 8 & - \\
			Grasping, Underwater, Homecare, Integrated & <15517 & <1015 & <9 & - \\
			%\hline
			%	POSys.(n) & $2^{n}$ & 3 & $2n+1$ & - \\
			%	Colli. Avoid. & 9720 & 18 & 3 & - \\
			%RS[11,11] & 247809 & 2 & 16 & $\checkmark$ \\
			%RS[15,15] & $7\times10^{6}$ & 3 & 20 & $\checkmark$ \\
			LaserTag & 4830 & $1.5\times10^{6}$ & 5 & - \\
			%	Light Dark & $\sim30$ & $\mathbb{R}$ & 5 & - \\
			%	Sub Hunt & $1.3\times10^{6}$ & $\mathbb{R}^{8}$ & 6 & $\checkmark$ \\
			%	V.D.P. Tag & $\mathbb{R}^{4}$ & $\mathbb{R}^{8}$ & $\mathbb{R}^{2}$ & - \\
			%\hline
			battleship & $\sim10^{18}$ & 2 & 100 & - \\ 
			Pocman & $\sim10^{56}$ & 1024 & 4 & $\checkmark$ \\
			%battleship, Pocman & <$\sim10^{56}$ & <1024 & <100 & $\checkmark$ \\
			Flickering Atari & $\sim10^{56}$ & $>10^{56}$ & $18$ & - \\
			Starcraft II & $10^{1685}$ & $>10^{1685}$ & $\sim10^{26}$ & $\checkmark$ \\
			\textbf{Mask Atari} & \textbf{$\sim10^{56}$} & \textbf{$>10^{56}$} & \textbf{$>90$} & \textbf{$\checkmark$} \\
			\hline
		\end{tabular}
		\label{tab:POMDPEnvs}
	\end{table}
	\subsubsection{POMDP Algorithms}
	The two ways to change a POMDP into a Markov decision process (MDP) are the information state approach and the belief state approach. For planning solvers, a belief state approach is used. Approximate methods such as point-based value iteration \cite{shani2013survey} restrict value function computations to a finite subset of the belief space, permitting only local value updates for this subset to avoid the exponential growth of the value function. This makes it applicable for domains with several thousands of states. For DRL, they correspond to encode histories by remembering features of the past \cite{mccallum1993overcoming} or by performing inference to determine the distribution over possible latent states \cite{kaelbling1998planning}. The deep recurrent Q-network (DRQN) \cite{hausknecht2015deep} trains an RNN to summarize history. \cite{igl2018deep} proposed an n-step A2C version of DRQN. 
	
	\section{Mask Atari}
	\subsection{Environment Features}
	Our new Mask Atari benchmark solves POMDP problems with DRL customized on Atari 2600 games \cite{brockman2016openai}. The core idea of Mask Atari is to introduce a controllable mask as the observable area of an agent to make the playing of Atari games to be a POMDP problem. The specifications of the mask operation are fully customizable through the provided interface, which consists of scale\&position, speed\&direction, quantity, and resolution. The details of each setting are as follows. 
	\begin{figure}%[ht]
		\centering
		\includegraphics[scale=0.45]{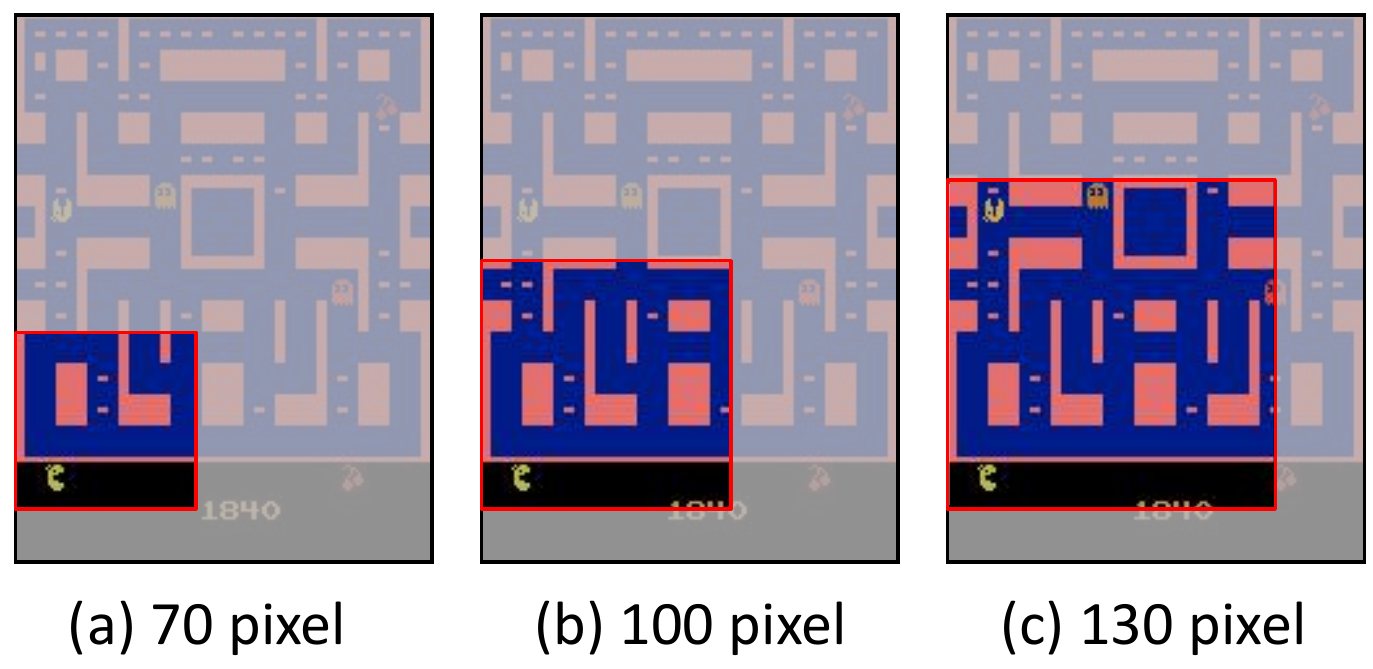}
		\caption{Scale setting for Mask Atari.}
		\label{fig:scale}
	\end{figure}
	\subsubsection{Mask Scale \& Position} 
	Let the game window size be ${L,W}$, where $L$ is the length of the window and $W$ is the width of the window. We define the scale of the mask as ${l,w}$, where $l \in (0,L)$ and $w \in (0,W)$. Fig.\ref{fig:scale} shows examples of different scale settings of masks in one Atari game. Because most sensors are isotropic, we recommend equalizing $l$ and $w$ as the standard setting. We provide two patterns for setting the initialized position of the mask, which is defined as the centric pixel coordinates in the mask: random position starting and centric position in the game window starting. The random position takes the boundary processing into consideration so the mask will be fully contained with the set scale in the game window. 
	\begin{figure}%[ht]
		\centering
		\includegraphics[scale=0.45]{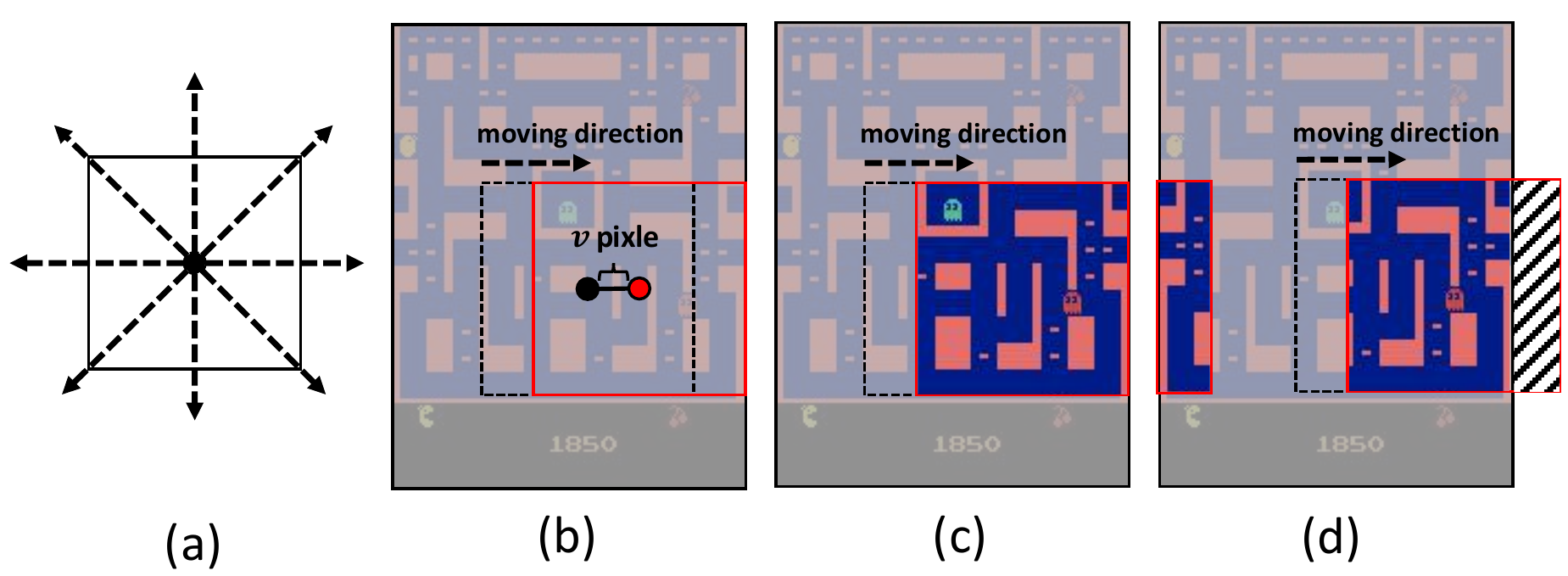}
		\caption{Direction\&Speed setting for the mask: a) 8-directions for the mask moving, b) example for mask moving with $v$ pixel, c) boundary stopping process when the mask reached the edge of the game window (the red box is where the mask stopped after moving), and d) boundary slipping through process.}
		\label{fig:speed}
	\end{figure}
	\begin{figure}%[ht]
		\centering
		\includegraphics[scale=0.45]{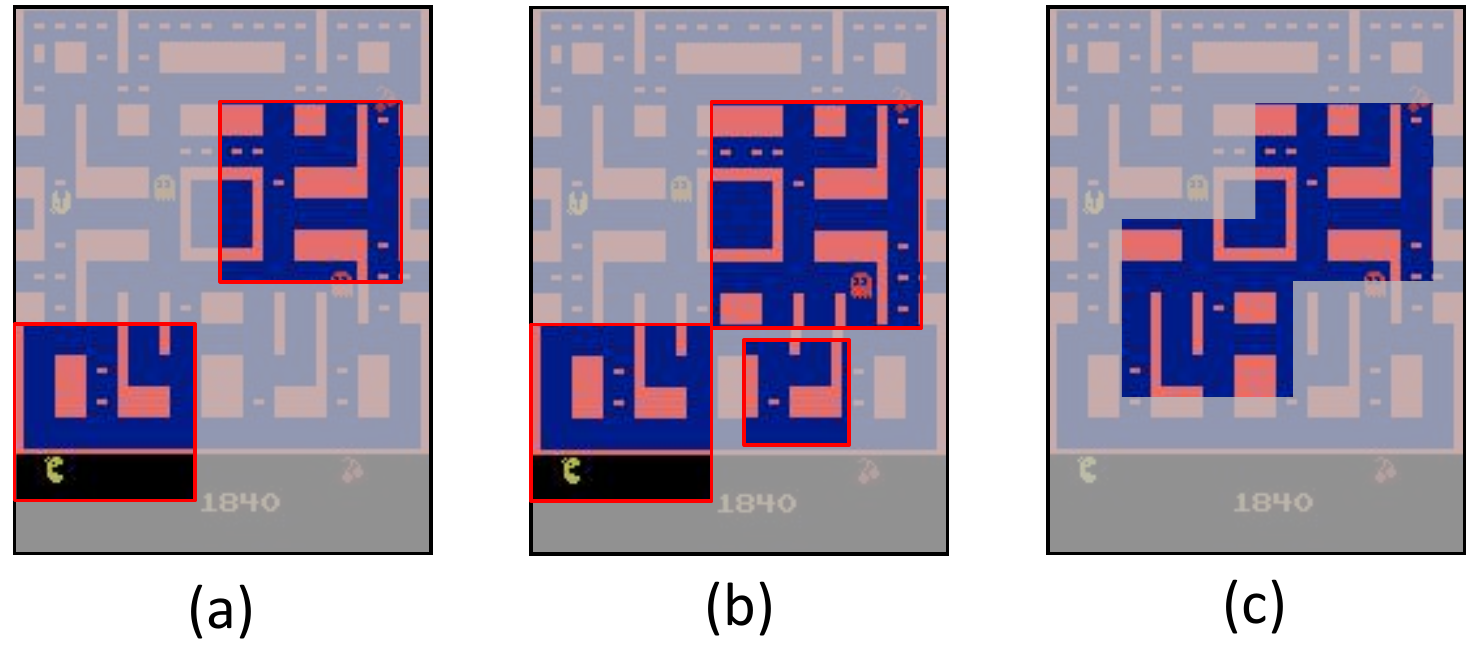}
		\caption{Examples of multiple masks in one game window: a) two masks without overlapping; only the area in the mask is observable, b) multiple masks can be set with different scales and moving speeds, c) multiple masks when overlapped with each other.}
		\label{fig:multiple}
	\end{figure}
	\subsubsection{Mask Speed \& Direction} 
	Mask Atari provides eight moving directions for the mask: \{left,right,up,down,left-up,left-down,right-up,right-down\}, as shown in Fig.\ref{fig:speed}. The moving speed of the mask is defined as $v$, where the mask can move $v$ pixels in one direction from the 4-directions \{left,right,up,down\} in each frame and moves $\lceil v/\sqrt{2} \rceil$ pixels on both the x-axis and y-axis for the \{left-up,left-down,right-up,right-down\} directions. When the mask reaches the edge of the game window in one movement, two kinds of boundary settings are provided, as shown in Fig. \ref{fig:speed}. $(1)$. “Boundary stopping” means the mask will be stopped upon reaching the boundary of the game window. $(2)$. “Boundary slipping through” means the mask will slip through the boundary and appear from the other side during the moving when reaching the boundary of the game window.
	
	\subsubsection{Multiple Masks} 
	Mask Atari also supports multiple masks. Each mask has its own action space, and the total number of action spaces will be a tensor product of each mask's action space. Moreover, each mask is independent of each other in our setting, so each mask also owns its scale and moving speed. When multiple masks overlap each other, the observable area for the agent will be the intersection area of the overlapped masks, as shown in Fig.\ref{fig:multiple}. 
	\begin{figure}%[ht]
		\centering
		\includegraphics[scale=0.4321]{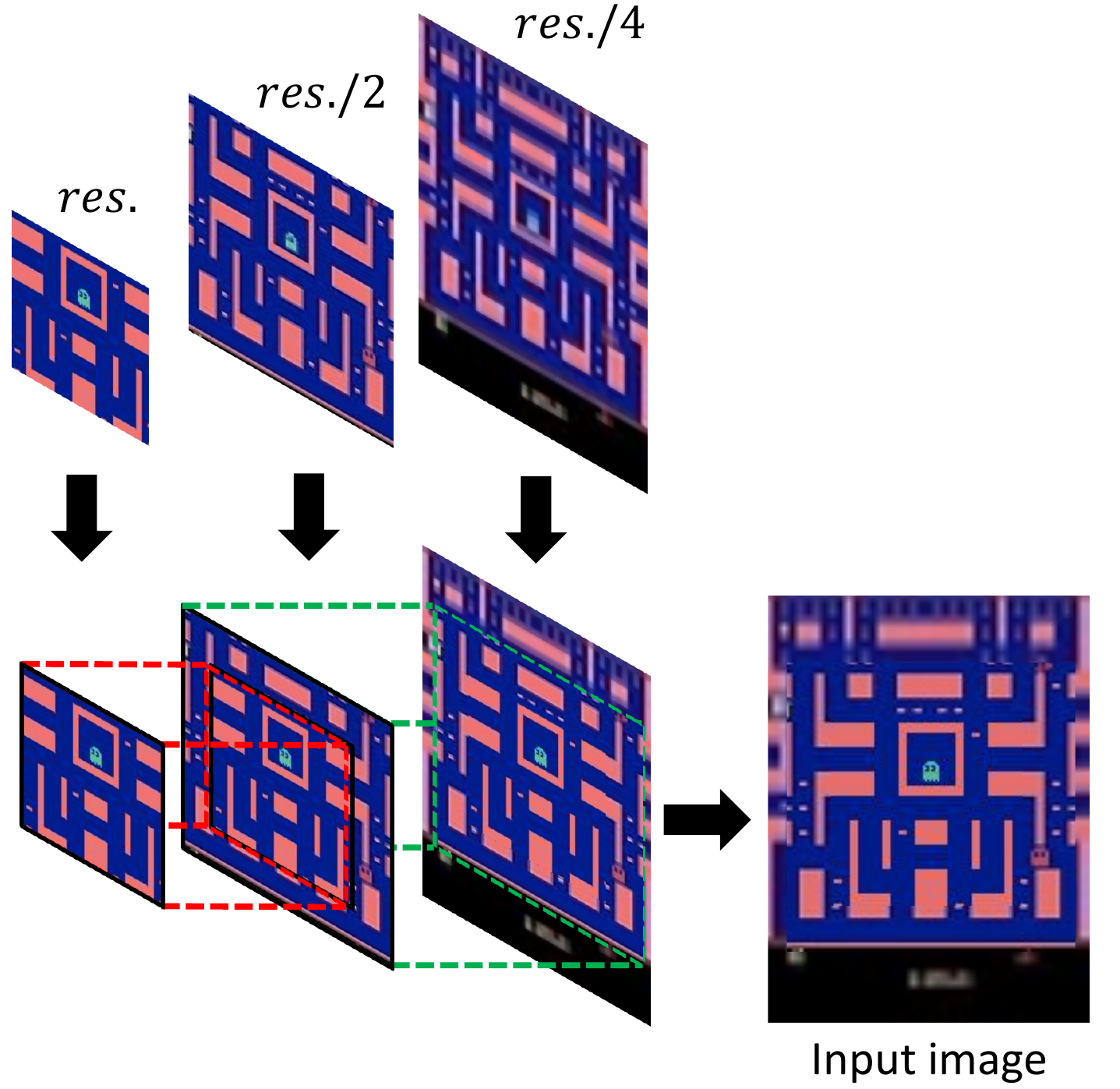}
		\caption{Mask-based resolution decay. The red box is the first layer and indicates the centric mask from the user setting. The resolution in the red box is 1x. The green box is the second layer that is 1.5x larger than the given mask scale located in the same centric position as the given mask. The resolution in the green box is 0.5x. The other area outside the green box is the third layer; its resolution is 0.25x. We use the composition of the three layers’ resolution under the corresponding area to generate the input image.}
		\label{fig:resolution}
	\end{figure}
	\subsubsection{Resolution Decay} 
	Our default setting provides full resolution in the mask, thus, in all observable areas. The default mask divides underlying observations into observable parts and unobservable parts, which simulate the field-of-view limitation. Inspired by foveated rendering \cite{guenter2012foveated} for simulating the focus area of the human vision system, we also provide a resolution decay option. Fig.\ref{fig:resolution} shows the sample of resolution decay in our environment. We set three layers for resolution decay; the first layer shown in Fig. \ref{fig:resolution} with the red box is the mask set by the user with 1x original resolution as $res_1$. The green box is 1.5x larger than the mask area and shares the same centric position with the mask that has 0.5x resolution as $res_2$. The other area outside the green box has 0.25x resolution as $res_3$. With this kind of setting, we can use the mask operation as a trial for introducing the receptive field in the human vision system into a simulation environment for an agent. That means the evaluations are not biased from the sensing ability and purely focus on the cognitive performance of the methods when compared with the human baseline.
	
	\subsection{Baseline Setting}
	\subsubsection{A2C-RNN Baselines}
	For POMDPs, some previous works in DRL have relied on training an RNN to summarize history information, such as the deep recurrent Q-network (DRQN) \cite{hausknecht2015deep}. While the original DRQN used Q-learning to train the policy, an n-step A2C version was proposed by \cite{igl2018deep}. Advantage actor critic (A2C) \cite{wu2017scalable} is a widely used DRL algorithm. It maintains a state-value function with a policy function to use the TD residual form of the policy gradient theorem, reducing the training variance while keeping the bias unchanged. A2C avoids drawing entire trajectories from a replay buffer and enables continuous actions. For our A2C-RNN baseline setting, we minimally modified the architecture of a deep Q-network (DQN) reported in \cite{mnih2015human}, replacing only its first fully connected layer with a recurrent LSTM layer of the same size as 512 and using A2C to train the policy. Meanwhile, we also report the result of original CNN architecture in the deep Q-network (DQN) \cite{mnih2015human} with A2C namely A2C-CNN in the evaluation part as reference.
	
	\section{Evaluation}
	\subsection{Evaluation Setting}
	\subsubsection{Game Setting}
	We tested ten Atari games that are suitable for A2C learning in a fully observable situation, four of which were also used by \cite{hausknecht2015deep} and \cite{igl2018deep} for Flickering Atari. For all the games, we used a frameskip of four and a stochastic version with a 0.25 chance of repeating the current action for a second time at each transition. We scaled the original $210\times160$ RGB frame to a $84\times84$ grayscale frame that follows the DQN \cite{mnih2015human}. The score reported in this section is the mean of the last 100 episode rewards after 10 million time steps. Each episode was started with 30 no operation actions. 
	
	\subsubsection{Mask Setting}
	We set the default parameters of our mask as a 27-inch screen at a 1-meter display distance situation. Therefore, the solid angle of the game window was around $20^{\circ}$. 
	A human eye can sense fine details only within a $5^{\circ}$ central circle. This tiny portion of the visual field projects to the retinal region called the “fovea,” which is tightly packed with colour cone receptors. The angular distance away from the central gaze direction is called “eccentricity.” Acuity falls off rapidly as eccentricity increases due to reduced receptor and ganglion density in the retina. The minimum discernible angular size (the reciprocal of visual acuity) increases roughly linearly with eccentricity \cite{strasburger2011peripheral}. \cite{guenter2012foveated} showed that rendering three nested and overlapping render targets or eccentricity layers centred around the current gaze point can create a human undetectable foveation image. These layers denote the inner/foveal layer, middle layer, and outer layer. The inner layer is smallest in angular diameter and rendered at the highest resolution (native display). The two peripheral layers cover a progressively larger angular diameter but are rendered at a progressively lower resolution. The layer's regions are found after optimizing against the minimum angle of resolution, which predicts acuity as a function of eccentricity. This optimization is sensitive to screen width and display distance. Regarding the display situation we supposed, we set our default mask size to 100 pixels to simulate the observation field up to the middle layer. 
	
	The peak angular speed of the human eye during a saccade reaches up to $900^{\circ}/s$. To track moving objects, achievable speed is associated with the saccade amplitude allowed. $30^{\circ}$ amplitude is associated with $500^{\circ}/s$, and $10^{\circ}$ is associated with $300^{\circ}/s$ \cite{britannica1987sensory}. Standard Atari games update at a frequency of 60 frames per second. Therefore, a reasonable speed for video gaming is about $5^{\circ}$ per frame. For our supposed $20^{\circ}$ display for the game window, around four frames are needed to complete the saccade. Therefore, we chose 50 pixels per frame as the default speed of the mask. 
	
	\subsection{Evaluation Methods}
	To investigate the performance of the fashion methods on our benchmark, we report the A2C-CNN as one baseline due to it being a widely used method in DRL for standard Atari games \cite{wu2017scalable}. We also report the results of A2C-RNN \cite{igl2018deep}, as it is a widely used method for Flickering Atari \cite{hausknecht2015deep}, which are also Atari-based POMDPs. Moreover, we report the results of using both methods without a mask as an observation limitation to their upper boundary performances. 
	\begin{enumerate}
		\item A2C-CNN(Full): We test the A2C-CNN algorithm model combination on the full observation setting so the whole game window can be observed. This test shows our implementation is comparable to the original implementation from \cite{wu2017scalable}. 
		\item A2C-RNN(Full): We also test the A2C-RNN algorithm model combination on the full observation setting because it is reportedly more suitable for POMDP environments \cite{hausknecht2015deep}. We use this result as the upper boundary performance of the test method to show the challenge of the masked condition. 
		\item A2C-CNN(Mask): We test the A2C-CNN algorithm model combination on our Mask Atari games as a reference for comparison. 
		\item A2C-RNN(Mask): We test the A2C-RNN algorithm model combination on our Mask Atari games to show the difficulty with the fashion method for POMDPs. 
	\end{enumerate}
	
	We used 64 parallel environments for the A2C algorithm's sampling. For all the games, we set the roll out number to 5, the frame stack to 4, the learning rate to 7e-4, the learning rate schedule to linear decay, the minimal learning rate to 7e-6, the optimizer to RMSprop, the epsilon to 1e-5, alpha to 0.99, the coefficient of the value loss function to 0.5, the maximum normalization of the gradient to 0.5, the coefficient of entropy to 0.01, and the discount factor gamma to 0.99. The same CNN neural network architecture with \cite{mnih2015human} is used as the policy and value approximate function model. The same RNN neural network architecture with \cite{hausknecht2015deep} is used. 
	\begin{figure*}%[!ht]%[!htb]
		\centering
		\includegraphics[width=3.5cm]{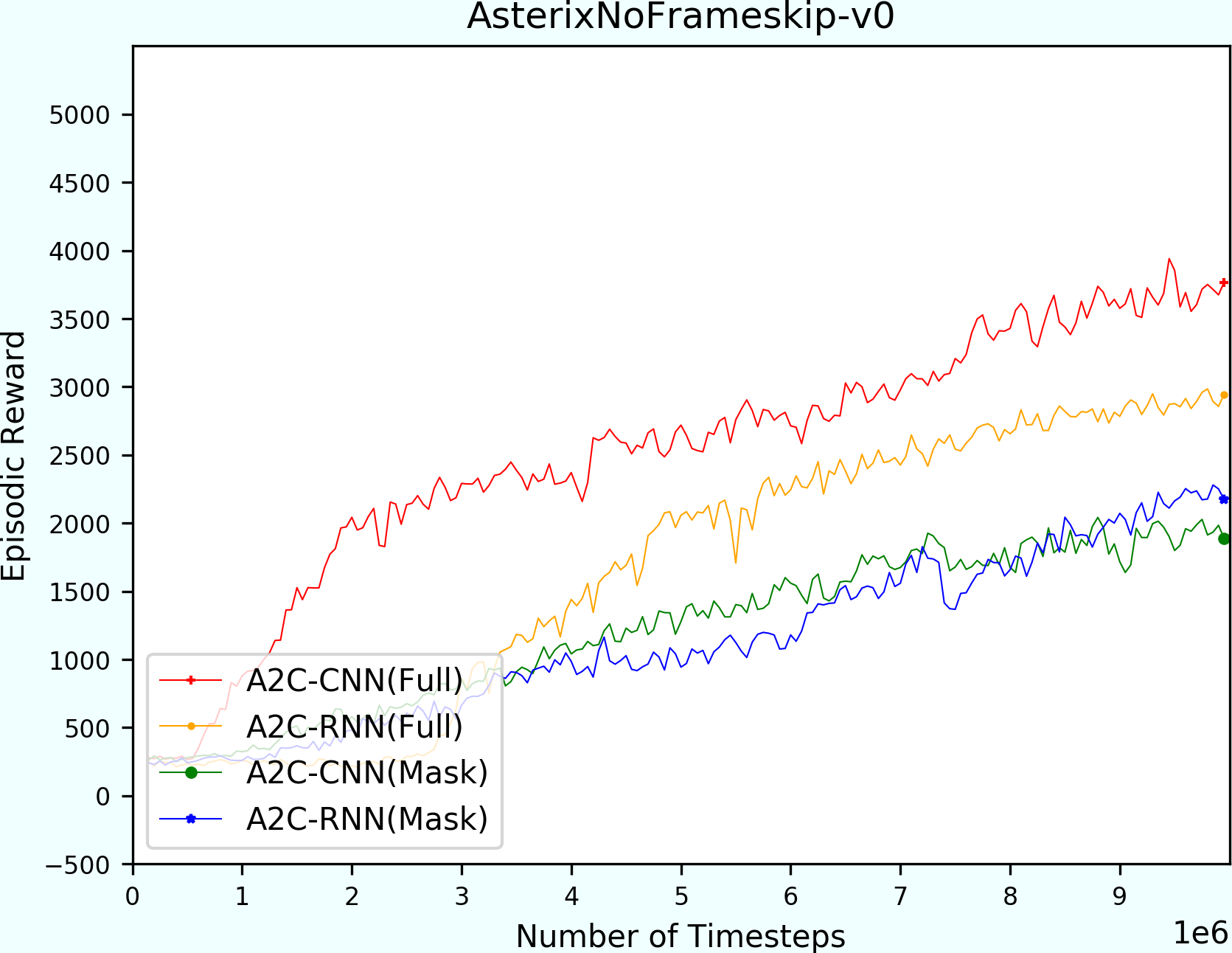}\hspace*{3pt}
		\includegraphics[width=3.5cm]{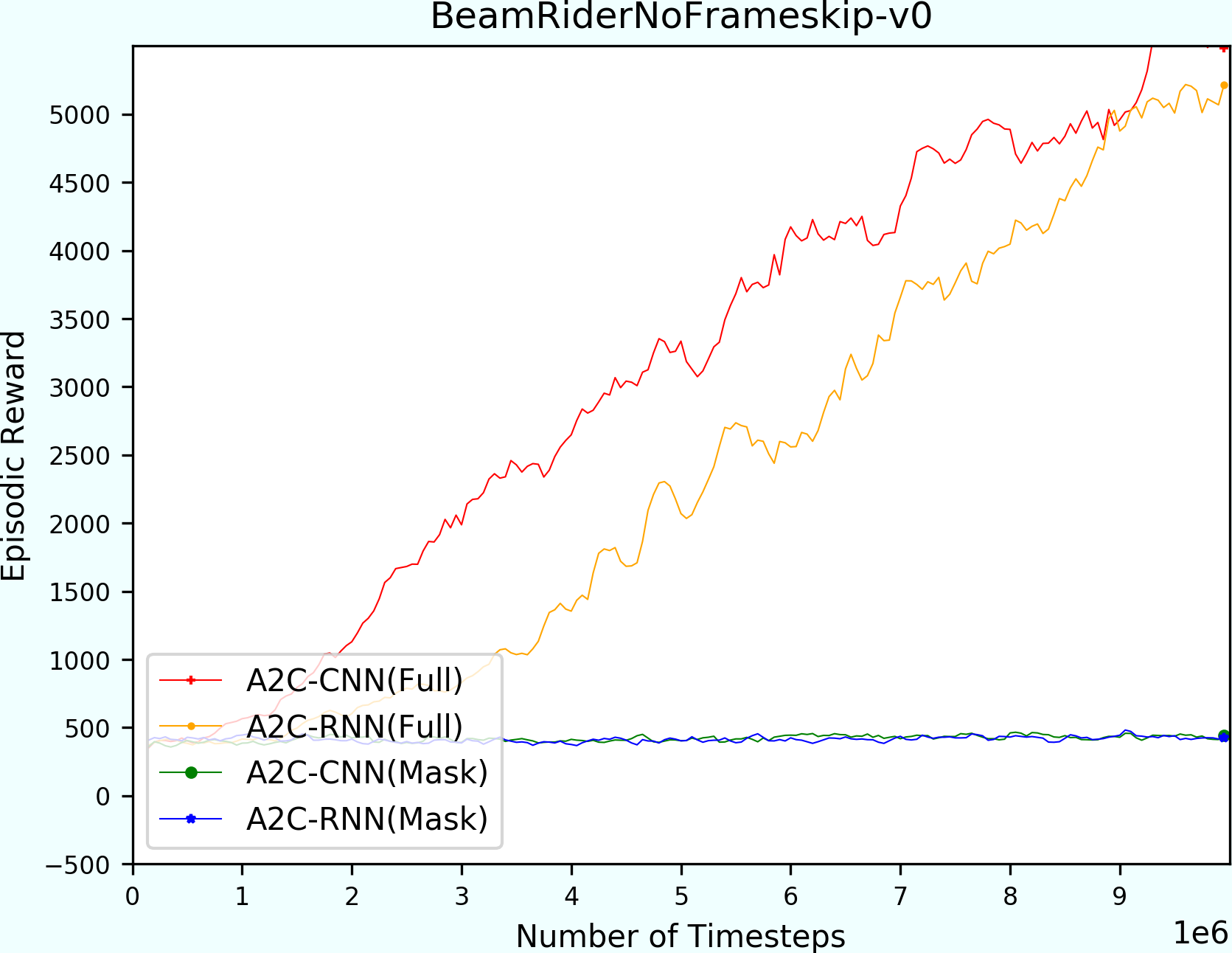}\hspace*{3pt}
		\includegraphics[width=3.5cm]{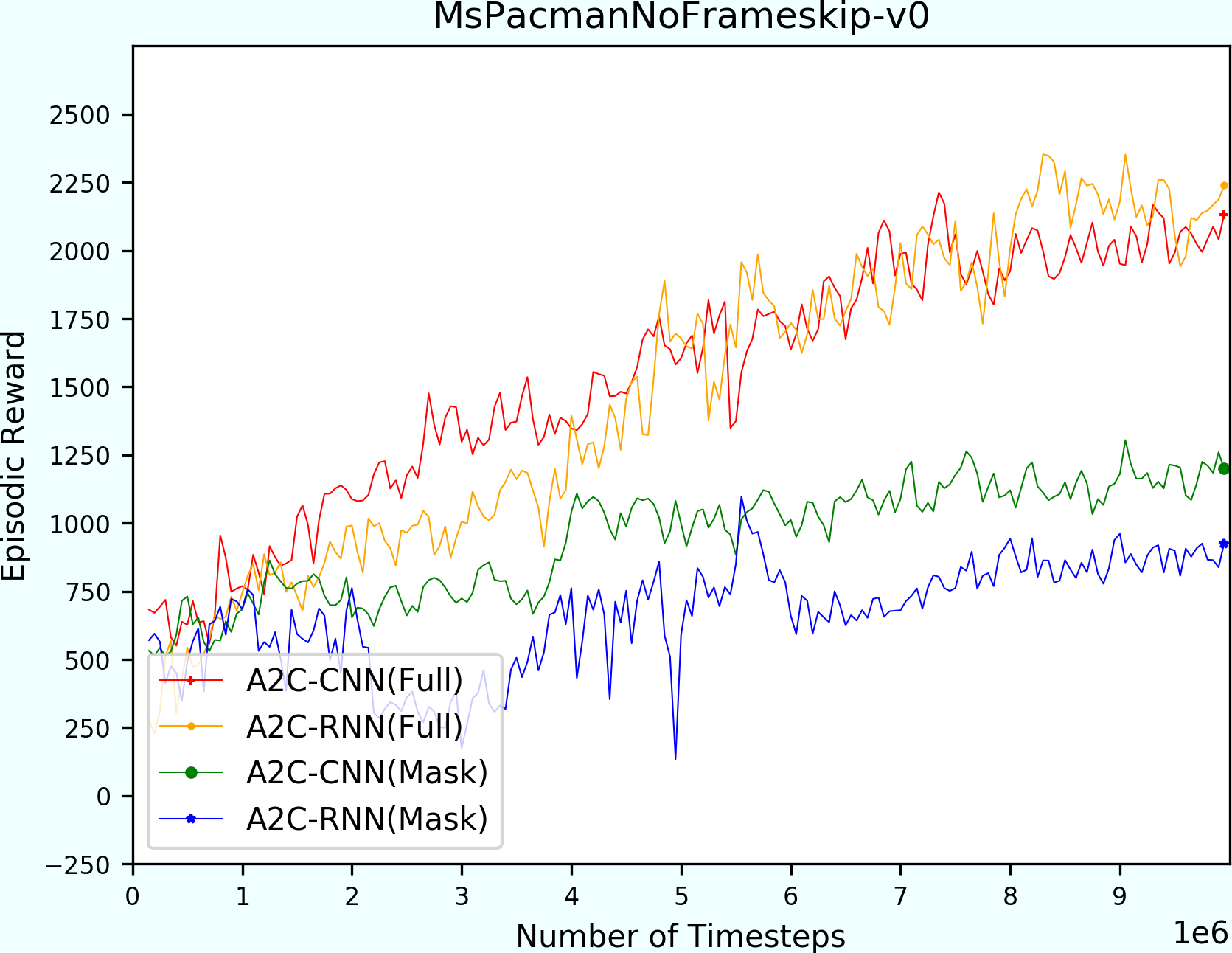}\hspace*{3pt}
		\caption{Learning curves of Asterix, BeamRider, and MsPacman.}
		\label{Curves}
	\end{figure*}
	\begin{table*}%[ht]
		\caption{10 M steps (40 M frames) training results.}
		\centering
		\begin{tabular}{lcccccc}
			\hline
			Environment & Human & A2C-RNN & A2C-RNN & A2C-CNN & A2C-CNN & Random \\
			& \scriptsize\cite{mnih2015human} & (Full) & (Mask) & (Full) & (Mask) & \scriptsize\cite{mnih2015human} \\
			\hline
			Assault & 742 & 3789 & 337 & 4428 & 495 & 222 \\
			Asterix & 8503 & 2943 & 2174 & 3765 & 1888 & 210 \\
			Asteroids & 47389 & 1786 & 1452 & 1844 & 1362 & 719 \\
			BeamRider & 16927 & 5215 & 454 & 5486 & 445 & 364 \\
			Breakout & 30.5 & 369 & 0.9 & 381 & 119 & 1.7 \\
			Centipede & 12017 & 4541 & 3252 & 3171 & 2665 & 2091 \\
			MsPacman & 6952 & 2239 & 837 & 2131 & 1200 & 307 \\
			Qbert & 13455 & 5173 & 187 & 9248 & 549 & 164 \\
			SpaceInvaders & 1669 & 1108 & 218 & 817 & 487 & 148 \\
			StarGunner & 10250 & 41989 & 1872 & 49395 & 1661 & 664 \\
			\hline
		\end{tabular}
		\label{tab:Baseline}
	\end{table*}
	\subsection{Baseline Performance}
	Table \ref{tab:Baseline} shows ten million steps training results of all environments. Fig.\ref{Curves} shows three example learning curves. Generally, an agent’s performance is substantially degraded with a limited field of view. In several cases in which long-range AIG was important, such as BeamRider and Qbert, the agent’s performance even dropped down to the same level of a random agent. 
	
	For BeamRider, the agent needs to control a spaceship to shoot moving enemy ships while avoiding being shot by the enemy ships. Unlike a full observation situation, when the field of view is limited, the agent needs to learn to move its field of view between enemy ships and its own ship. Otherwise, when it focuses on its own ship, enemy ships will be kept outside its field of view, and only the enemy’s bullets can be seen. This situation makes it more difficult for the agent to dodge bullets. For Qbert, the agent needs to work through all the steps of a pyramid while avoiding being caught by enemies. When close to clearing the stage, the remaining steps are always outside the agent’s field of view. Therefore, it is important for the agent to learn to temporarily move its focus away from itself to actively search where the steps have not arrived. 
	
	\subsection{Ablation Study}
	To investigate the effect of the hyperparameter settings of the masks, we did three ablation studies on four games that were also used by \cite{hausknecht2015deep} and \cite{igl2018deep} for scale, speed, and numbers. 
	\begin{table}%[ht]
		\caption{Ablation study}
		\begin{subtable}{.34\textwidth}
			\centering
			\begin{tabular}{lccc}
				\hline
				Scale & 70 & 100 & 130 \\
				\hline
				Asteroids & 1039 & 1990 & 2522 \\
				BeamRider & 430 & 427 & 1498 \\
				Centipede & 3308 & 2955 & 3939 \\
				MsPacman & 451 & 926 & 1588 \\
				\hline
			\end{tabular}
			\label{subtab:Scale}
			\captionsetup{width=.8\linewidth}
			\caption{Ablation study for mask scale in pixel (mask speed: 50 pixels/frame, mask number: 1).}
		\end{subtable}
		\begin{subtable}{.34\textwidth}
			\centering 
			\begin{tabular}{lccc}
				\hline
				Speed & 10 & 30 & 50 \\
				\hline
				Asteroids & 1452 & 1556 & 1990 \\
				BeamRider & 454 & 413 & 427 \\
				Centipede & 3252 & 2834 & 2955 \\
				MsPacman & 837 & 1241 & 926 \\
				\hline
			\end{tabular}
			\label{subtab:Speed}
			\captionsetup{width=.8\linewidth}
			\caption{Ablation study for mask speed in pixel/frame (mask scale: 100 pixels, mask number: 1 ).}
		\end{subtable}
		\begin{subtable}{.3\textwidth}
			\centering 
			\begin{tabular}{lccc}
				\hline
				No. of Mask & 1 & 2 & \\
				\hline
				Asteroids & 1990 & 1587 & \\
				BeamRider & 427 & 423 & \\
				Centipede & 2955 & 3322 & \\
				MsPacman & 926 & 1390 & \\
				\hline
			\end{tabular}
			\label{subtab:Quantity}
			\captionsetup{width=.9\linewidth}
			\caption{Ablation study for mask quantity (mask scale: 100 pixels, mask speed: 50 pixels/frame).}
		\end{subtable}
		\label{tab:Ablationstudy}
	\end{table}
	\begin{figure*}%[!ht]
		\centering
		\includegraphics[width=1.86cm]{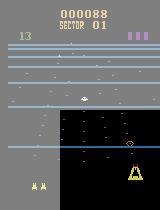}\hspace*{1pt}
		\includegraphics[width=1.86cm]{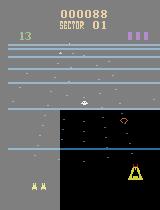}\hspace*{1pt}
		\includegraphics[width=1.86cm]{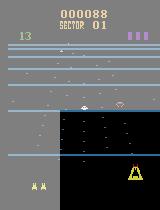}\hspace*{1pt}
		\includegraphics[width=1.86cm]{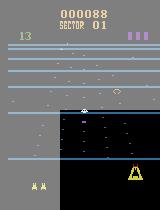}\hspace*{1pt}
		\includegraphics[width=1.86cm]{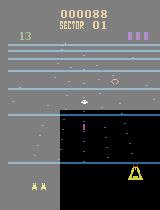}\hspace*{1pt}
		\includegraphics[width=1.86cm]{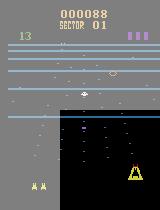}\hspace*{1pt}
		\caption{Enemy ship shooting and running in BeamRider (Mask scale = 100).}
		\label{Screenshot}
	\end{figure*}
	\subsubsection{Mask Scale}
	To investigate the effect of scaling the mask, we fixed the moving speed of the mask to a default value of 50 pixels per frame and changed the scale of the mask from 70 pixels to 130 pixels. Table \ref{tab:Ablationstudy}(a) shows the results. Although a small-scaled mask generally makes games more challenging, mask scaling has different effects on different games. For games that need long-range AIG, such as BeamRider, the default value of the mask scale, 100 pixels, pulls down the scores close to that of a random agent. However, original challenge tasks and long-range information gathering is not quite critical for games such as Asteroids, so the default mask scale has a limited effect on the scores. 
	
	BeamRider's score drops rapidly when the mask scale changes from 130 pixels to 100 pixels because the enemy ship always shoots from a distance of approximately 100 pixels. Fig.\ref{Screenshot} shows a screenshot of this situation. The gray parts represent how the mask limits the field of view. A small white enemy ship at the centre of the screenshot came from far away, shot its pink bullet near the edge of the mask, and left. Therefore, with a mask scale of 100 pixels, the agent focused on itself cannot see the enemy ship, increasing the difficulty of dodging enemy bullets. When the mask scale equals 130 pixels, the agent does not need to move its focus between itself and the enemy ships, and can see them in one field of view. It dramatically changes the learning difficulty. 
	
	\subsubsection{Mask Speed}
	To investigate how moving speed changes the performance, we set the scale of mask to the default value of 100 pixels and changed the moving speed of the mask from ten pixels per frame to 50 pixels per frame. Table \ref{tab:Ablationstudy}(b) shows the results. Although a slow-moving mask generally makes games more challenging, the effect of changing the mask speed is relatively gentler than the effect of changing the mask scale. 
	
	\subsubsection{Multiple Masks}
	To investigate the effect of changing the number of masks, we set the scale of masks to the default of 100 pixels, set the moving speed of the masks to the default of 50 pixels per frame, and increased the number of masks to two. Table \ref{tab:Ablationstudy}(c) shows the results. The results show that half the games achieved a higher score while the other half achieved a lower score. We suspect that, while the increased total field of view increases the information that the agent can access, the increased total action space also increases the reinforcement learning difficulty. The final scores were affected by both the positive factors and the negative factors. 
	
	\subsection{Heuristic Ideas for AIG}
	Mask Atari provides a controllable mask, that makes the agent can use it for simulating the active information gathering (AIG) process. Thus, one typical usage is to accelerate the agent’s learning process for POMDPs by using the observed information in the mask to perform AIG. We propose two heuristic ideas we tested on the Mask Atari. Both ideas helped the agent learn faster by giving an auxiliary reward based on the information-gathering performance. We use these experiments to show that the internal structure of the AIG setting can be used to accelerate the learning process. However, achieving a more optimal policy than the baseline for complex, objective, vision-rich tasks like the Mask Atari is not easy. 
	\begin{enumerate}
		\item Seeking new information: We added an auxiliary reward $r=-0.25*(v_{t-3}+v_{t-2}+v_{t-1}+v_{t}) \cdot v_{t+1}$ to the original game reward. Here, $v_{i}$ is a normalized vector calculated by flattening the observed image in time $i$. This auxiliary reward punishes the agent when it observes the same image with the previous average of the stacked observed image, inspiring the agent to get more new information. 
		\item Expanding the field of view: We added an auxiliary reward $r=\left\|w_{t-3,t+1}-w_{t-3,t}\right\|$ to the original game reward. Here, $w_{i,j}$ is a vector calculated by flattening the merged mask image from time $i$ to time $j$. A merged mask image is an array that equals 1 if the corresponding pixel is covered by a mask from time $i$ to time $j$ and zero otherwise. 
	\end{enumerate}
	
	\begin{figure*}[t]
		\centering
		\includegraphics[width=3.5cm]{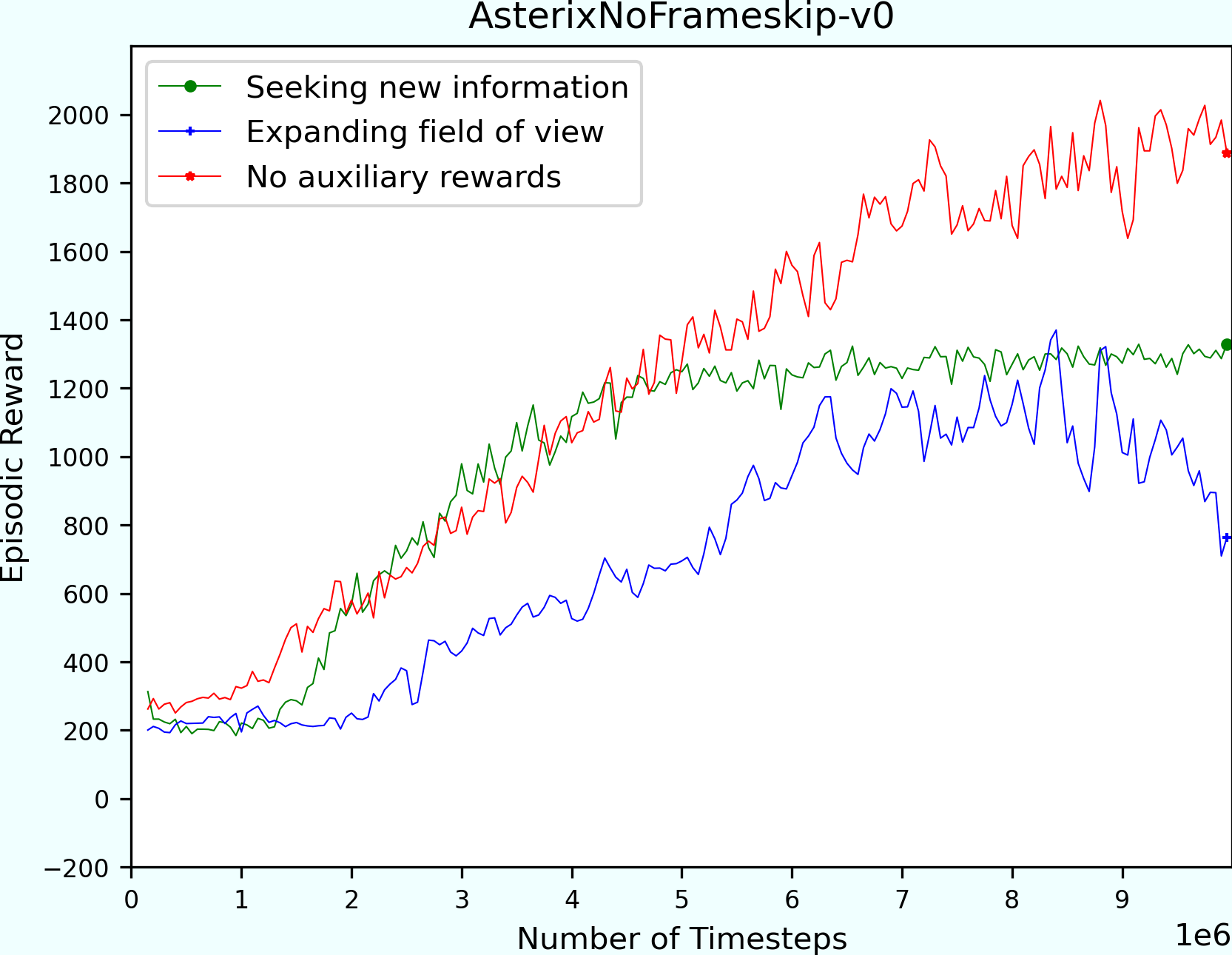}\hspace*{3pt}	\includegraphics[width=3.5cm]{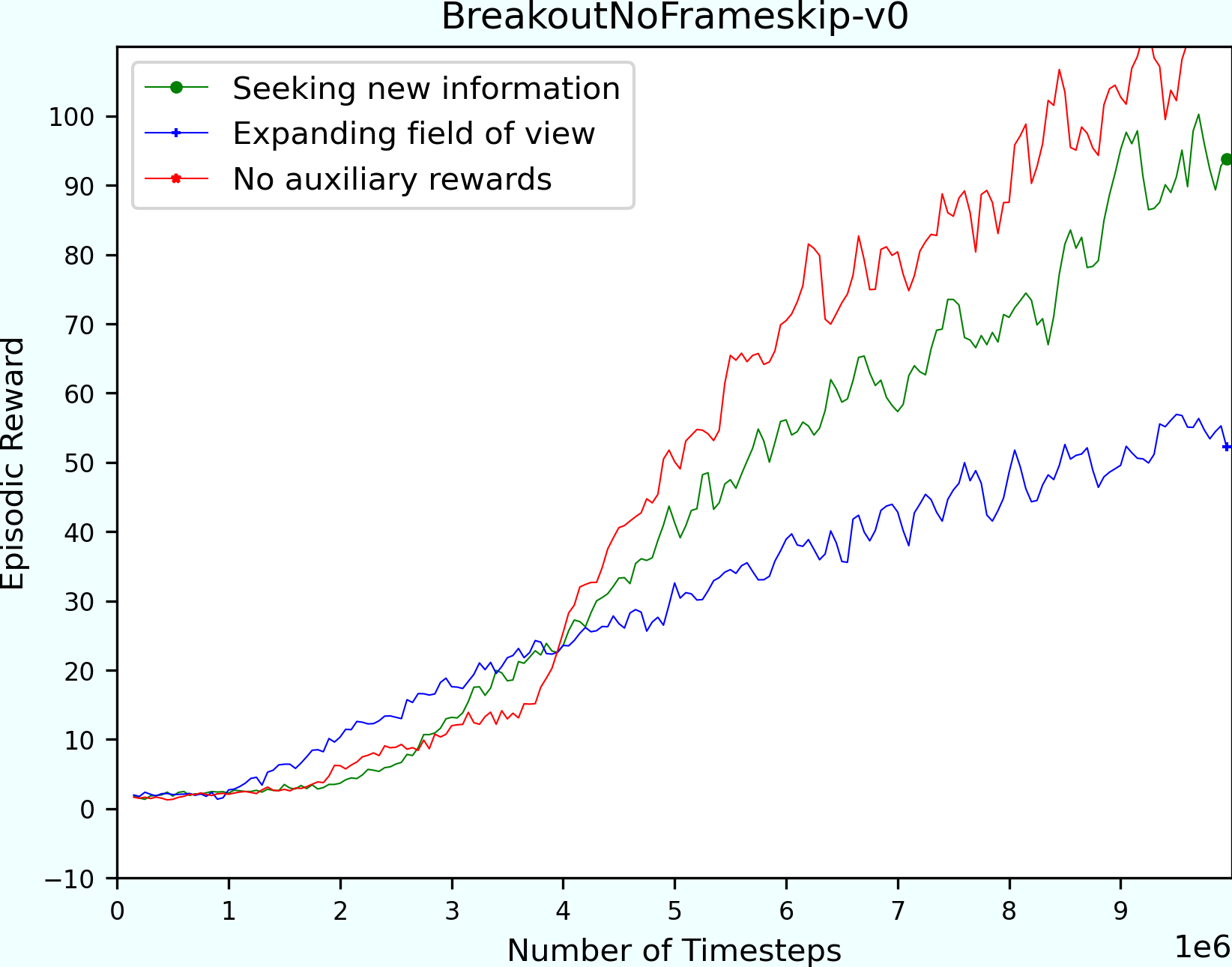}\hspace*{3pt}
		\includegraphics[width=3.5cm]{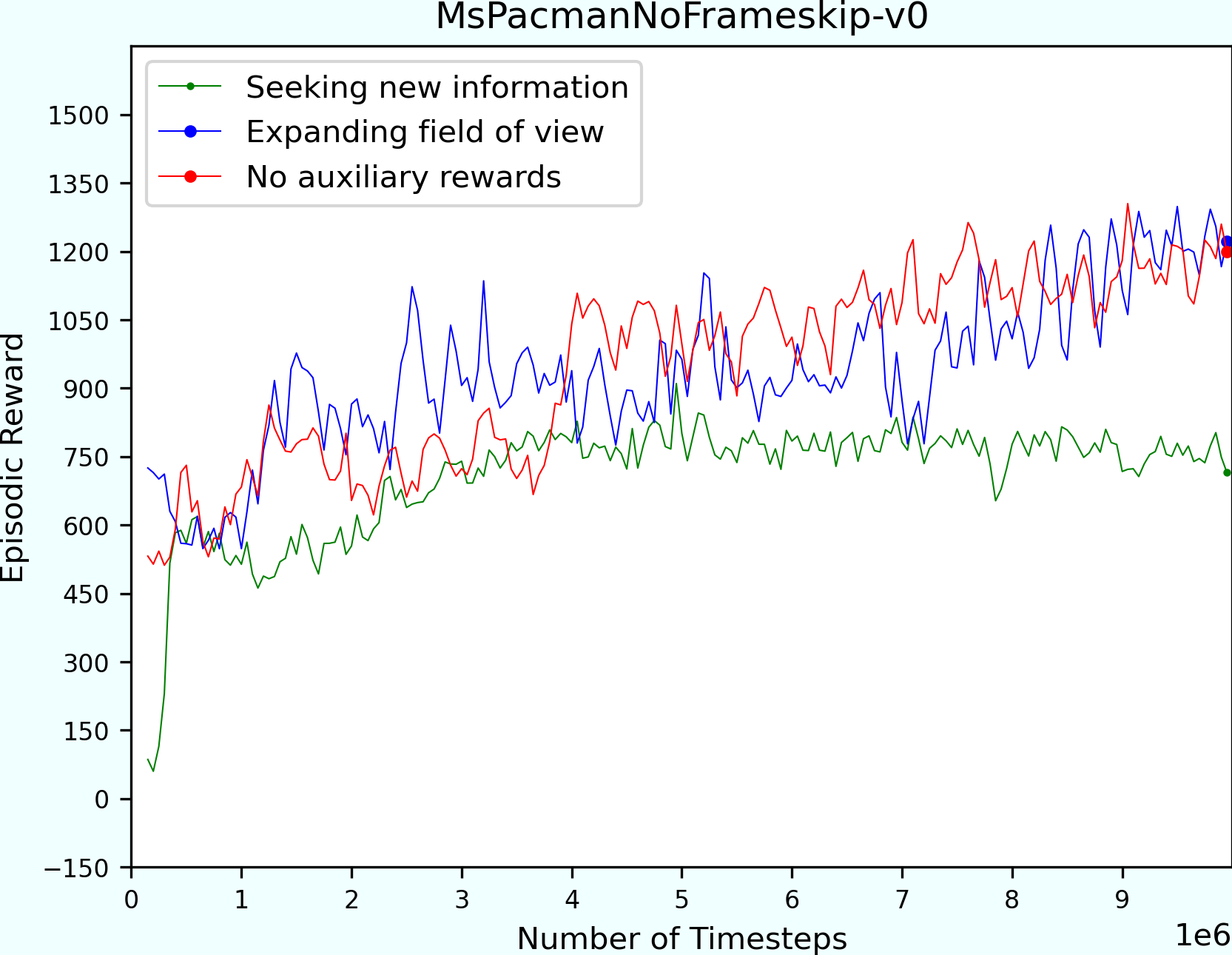}\hspace*{3pt}
		\caption{Learning curves with auxiliary rewards.}
		\label{Auxiliary}
	\end{figure*}
	
	Fig.\ref{Auxiliary} shows the results of the two heuristic ideas compared with the original, no-auxiliary-rewards version with Asterix, Breakout, and MsPacman. Method A2C-CNN is used. All mask parameters are set to their defaults, meaning one mask with a 100-pixel scale and a 50-pixel-per-frame speed. Auxiliary rewards successfully accelerated the learning process of the agent, but also led the final policy to a sub-optimal result. 
	
	\section{Conclusion}
	We presented Mask Atari, a new benchmark for solving POMDP problems with AIG by DRL-based approaches. Mask Atari was constructed based on Atari 2600 games with a controllable, moveable, and learnable mask as the observation area for the target agent. The masks are fully customizable through the provided interface. Mask Atari provides a challenging, efficient benchmark for AIG in complex, objective, vision-rich tasks. Its operation is a trial for introducing the receptive field in the human vision system into a simulation environment for an agent. Therefore, the evaluations are not biased from the sensing ability and purely focus on the cognitive performance of the methods when compared with the human baseline. 
	We evaluated several baseline algorithm and model combinations with Mask Atari with an ablation study on different settings of the masks to show the current popular DRL based method performance on it.
	To further utilize the feature of Mask Atari, we proposed two heuristic ideas and tested them to show the opportunities and challenges of the hand-eye coordination required by Mask Atari. Both of them helped the agent learn faster by giving an auxiliary reward based on the information-gathering performance. The auxiliary rewards successfully accelerated the learning process for the agent but also led the final policy to a sub-optimal result. That means the internal structure of the AIG setting can be used to accelerate the learning process but achieving a more optimal policy than the baseline for complex, objective, vision-rich tasks like Mask Atari is not easy. More complicated methods such as assembled auxiliary rewards, fade away auxiliary rewards, factorized models, and Bayesian optimization will be considered and tested in future work. 
	
	\bibliographystyle{splncs04}
	\bibliography{samplepaper}
	
\end{document}